\documentclass{article}
\usepackage{ijcai24}

\usepackage{times}
\usepackage{soul}
\usepackage{url}
\usepackage[hidelinks]{hyperref}
\usepackage[utf8]{inputenc}
\usepackage[small]{caption}
\usepackage{graphicx}
\usepackage{amsmath}
\usepackage{amsthm}
\usepackage{booktabs}
\usepackage{algorithm}
\usepackage{algorithmic}
\usepackage[switch]{lineno}
\usepackage{color} 

\usepackage{multirow}

\usepackage{amssymb}
\usepackage{arydshln}  

\usepackage{hyperref}

\hypersetup{hidelinks,
	colorlinks=true,
    allcolors=black,    
	urlcolor=blue,
	pdfstartview=Fit,
	breaklinks=true}

\title{Improving Pseudo Labels with Global-Local Denoising Framework\\for Cross-lingual Named Entity Recognition}

\author{
Zhuojun Ding$^{1,2}$
\and
Wei Wei\thanks{Corresponding author}$^{1,2}$\and
Xiaoye Qu$^{1,2}$\and
Dangyang Chen$^{3}$
\affiliations
$^1$Cognitive Computing and Intelligent Information Processing (CCIIP) Laboratory, School of Computer Science and Technology, Huazhong University of Science and Technology\\
$^2$Joint Laboratory of HUST and Pingan Property \& Casualty Research (HPL)\\
$^3$Ping An Property \& Casualty Insurance Company of China, Ltd.
\emails
\{dingzj,weiw,quxiaoye\}@hust.edu.cn,
chendangyang273@pingan.com.cn
}

\begin{document}

\maketitle

\begin{abstract}
Cross-lingual named entity recognition (NER) aims to train an NER model for the target language leveraging only labeled source language data and unlabeled target language data. Prior approaches either perform label projection on translated source language data or employ a source model to assign pseudo labels for target language data and train a target model on these pseudo-labeled data to generalize to the target language. However, these automatic labeling procedures inevitably introduce noisy labels, thus leading to a performance drop.
In this paper, we propose a Global-Local Denoising framework (GLoDe) for cross-lingual NER. Specifically, GLoDe introduces a progressive denoising strategy to rectify incorrect pseudo labels by leveraging both global and local distribution information in the semantic space. The refined pseudo-labeled target language data significantly improves the model's generalization ability. {Moreover, previous methods only consider improving the model with language-agnostic features, however, we argue that target language-specific features are also important and should never be ignored}. To this end, we employ a simple auxiliary task to achieve this goal.
Experimental results on two {benchmark datasets} with six target languages demonstrate that our proposed GLoDe significantly outperforms current state-of-the-art methods.
\footnote{Our code is available at \href{https://github.com/Ding-ZJ/GLoDe}{https://github.com/Ding-ZJ/GLoDe}}
\end{abstract}

\section{Introduction}

Named Entity Recognition (NER) aims to identify spans belonging to predefined entity classes in a given text, such as locations and organizations~\cite{qu2023distantly,gu2022delving,qu2023survey}, which is an important information extraction task~\cite{zhu2023mirror,fan2023improving}. Deep neural networks trained on large amounts of labeled data have achieved excellent performance~\cite{wei2021position} on NER. However, most low-resource languages {(e.g. Albanian and Finnish)} still suffer from insufficient labeled data compared to high-resource languages, such as English. Moreover, annotating entities in texts is time-consuming and expensive. Such predicament motivates research on cross-lingual NER~\cite{xie2018neural}. Cross-lingual NER aims to leverage labeled data in a high-resource language (source language) and unlabeled data in a low-resource language (target language) to train an NER model, which generalizes well to the target language.

Previous solutions for cross-lingual NER can be divided into three categories: (1) Feature-based methods, aiming to design and learn language-agnostic features to improve generalization ability in the target language~\cite{wu2019surprising}. (2) Translation-based methods, conducting translation and label projection on source language data to obtain labeled target language data~\cite{jain2019entity}. (3) Teacher-student learning frameworks~\cite{hinton2015distilling}. These methods use a model trained on source language data to assign soft pseudo labels for target language data~\cite{ma2022wider}. Then, the pseudo-labeled data is used to train a target model.

\begin{figure}[!t]
	\centering
    \includegraphics[width=1\linewidth]{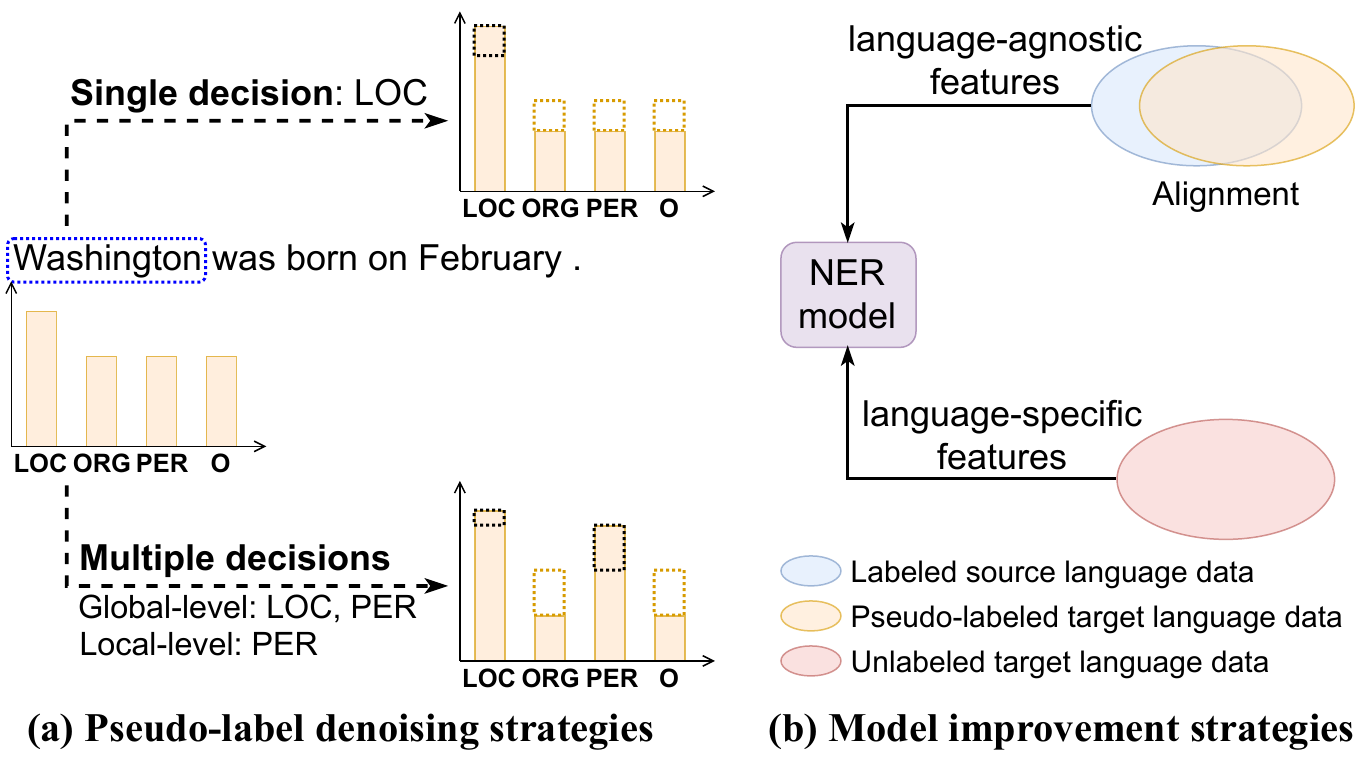}
	\caption{(a) \textbf{Up}: Previous works only consider one potential entity type for pseudo-label denoising. \textbf{Down}: We leverage global and local information within the semantic space to obtain multiple denoising decisions. (b) \textbf{Up}: Previous works only consider language-agnostic features to improve the NER model. \textbf{Down}: We leverage target language-specific features for model improvement.}
	\label{fig:model_intro}
\end{figure}

Recently, teacher-student learning frameworks have gradually become the mainstream solution. For these methods, the performance of the target model improves as the quality of pseudo labels increases. To this end, many improvement strategies have been proposed. 
Certain methods select high-confidence pseudo-labeled data for training through adversarial discriminators~\cite{chen2021advpicker} or reinforcement learning~\cite{liang2021rikd}. However, such data selection results in an increased distributional difference and fails to fully utilize target language data. 
There are also methods integrating multiple source models~\cite{wu2020tsl,wu2021unitrans} to generate pseudo labels. However, constructing various data sources to obtain diverse models significantly increases the training costs. 


Considering these aforementioned issues, we propose a \textbf{G}lobal-\textbf{Lo}cal \textbf{De}noising Framework (\textbf{GLoDe}).
Specifically, we leverage the global and local distribution information of the semantic space to denoise pseudo-labeled target language data. With our framework, high-quality data can be obtained without data selection or model ensemble. 
In detail, at the global level, we propose a prototype-based pseudo-label refinement method. Prototypes can be considered cluster centroids of each type. Conventionally, if a sample is most similar to a certain type's prototype, it is generally regarded as belonging to that specific type. 
~\cite{zhou2023conproto} initially leverages this strategy to iteratively improve pseudo labels.
However, misclassified samples may exhibit a higher degree of similarity to the prototype of the incorrect type rather than to prototypes of other types.
Therefore, we adopt a multiple-decision scheme to tackle it, {as shown in Figure~\ref{fig:model_intro}(a)}. 
When the similarity between a sample and the prototype of a particular type exceeds a specific threshold, the type is considered as a potential classification for that sample. 
{To affirm the credibility of similarity thresholds, we adopt dynamic thresholds for each entity type.} 
Consequently, a sample may have more than one potential type, which mitigates the problem of misclassification. 
At the local level, we leverage the distribution information of the sample's neighborhood. For a given sample, we retrieve its $K$-nearest neighbors in the semantic space. Based on the distribution of the neighbors' types, we define the similarity of the sample to each type. 
{Then, we compare these similarities with class-specific thresholds to obtain a potential classification for that sample, following the same procedure applied at the global level.} 
By integrating the decision from both global and local levels, we obtain holistic denoising directions for the pseudo label of samples. 

Additionally, prior methods typically improve the cross-lingual NER with language-agnostic features, adopting alignment tasks such as adversarial training or contrastive learning~\cite{zhou2023conproto}, as shown in Figure~\ref{fig:model_intro}(b). However, target language-specific features are never considered but are also beneficial. Therefore, we employ the Masked Language Model (MLM) task, which necessitates solely unlabeled target language data. 
This simple auxiliary task further enhances the model's understanding of the target language.

To sum up, our contributions can be concluded as follows: 

\begin{itemize}

\item {We design a global-local denoising framework that comprehensively leverages information within the semantic space to rectify the incorrect pseudo labels. A multiple-decision scheme with dynamic class-specific thresholds is devised to mitigate the problem of misclassification and significantly improve the quality of pseudo labels.}

\item {We propose to improve the cross-lingual NER with target language-specific features, which are never considered by previous work. To this end, we devise an auxiliary MLM task, which is both simple and beneficial.}

\item Experimental results on two benchmark datasets with six target languages demonstrate the superior performance of our approach. 

\end{itemize}

\section{Related Work}

Existing methods for cross-lingual NER mainly fall into three categories: feature-based methods, translation-based methods, and teacher-student learning methods.

\paragraph{Feature-based Methods}
These methods learn language-agnostic features so that the model trained on the source language data can better adapt to the target language. Earlier work leverages gazetteers~\cite{zirikly2015cross} and Wikifier features~\cite{tsai2016wikification}. The introduction of multilingual pre-trained language models has significantly improved the cross-lingual representations~\cite{conneau2020unsupervised,pires2019multilingual}. Some methods incorporate semantic alignment for further improvement through adversarial training~\cite{keung2019adversarial} and contrastive learning~\cite{wu2020explicit}. However, coarse-grained alignment is useless for NER. Entity-level alignment is advantageous, but the lack of ground truth labels in target language data inevitably leads to incorrect alignment. Furthermore, these methods do not acquire target language-specific features.

\paragraph{Translation-based Methods}
These methods translate the source language data and perform label projection to acquire labeled data in the target language. Some approaches employ word-to-word~\cite{xie2018neural} or phrase-to-phrase translation~\cite{mayhew2017cheap}, with entity labels directly linked to the translated content. However, substantial semantic information is sacrificed during translation. Therefore, other approaches consider translating the entire sentence~\cite{jain2019entity}, but the label projection process will introduce noise labels. \cite{liu2021mulda} introduces a placeholder-based approach that combines the aforementioned two strategies. \cite{yang2022crop} adopts the back-translation scheme. Nevertheless, the processes of translation and label projection inevitably carry the risk of introducing noise.

\paragraph{Teacher-student Learning Methods}
These methods leverage a source model (e.g., a model trained on source language data) to assign soft pseudo labels for target language data, subsequently utilizing this data to train a new model~\cite{wu2020tsl}. Soft labels offer more information and reduce the risk of overconfidence compared to hard labels~\cite{hinton2015distilling}. However, this remains insufficient. The key question is the quality of pseudo labels. \cite{ma2022wider,ge2023prokd} leverage the concept of feature-based methods to train the source model, aiming to generate higher-quality pseudo labels. \cite{chen2021advpicker,liang2021rikd} attempt to select a subset of the pseudo-labeled data with high confidence. However, data selection prevents the model from fully learning the characteristics of the target language data. Introducing multiple models~\cite{li2022unsupervised} is also widely used. \cite{wu2021unitrans,zhao2022transadv} integrate the predictions of multiple models as the final pseudo labels. \cite{zeng2022dualner,ma-etal-2023-colada} allow multiple models to generate pseudo labels for each other separately. However, various data sources are needed to train diverse models. For instance, \cite{zhao2022transadv} introduces source language data, translation data, and data with disrupted word order. 




\section{Methodology}

In this section, we describe our proposed framework GLoDe.
{Figure~\ref{fig:model_main}(a) presents the overall architecture of our approach.}

\subsection{{Overall Framework}}

\subsubsection{{Span-based NER Task}}

Following previous works~\cite{fu2021spanner,zhou2023conproto}, we formulate NER as a span classification problem. Formally, given a sentence $X=\{x_1,x_2,\dots,x_n\}$, our objective is to assign entity labels to each possible span $s_i=\{x_{b_i},x_{b_i+1},\dots,x_{e_i}\}$ in $X$, where $x_{b_i}$ and $x_{e_i}$ denote the beginning and end tokens of $s_i$.

For an input sentence $X$, we first obtain its token representations through a pre-trained language model (PLM): $H = \{h_i\}_{i=1}^n = \text{PLM}(X)$. Let $S(X)=\{s_i\}_{i=1}^m$ denote all possible spans in $X$. For each $s_i \in S(X)$, we introduce additional span length embedding $l_i$ and span morphology embedding $m_i$, which are both learnable embedding parameters. Then, we formulate the span representation by integrating the representations of the beginning and end tokens, the length embedding, and the morphology embedding: $z_i=[h_{b_i};h_{e_i};l_i;m_i]$.
Finally, $z_i$ is fed into a linear classifier to yield its probability distribution $p_i \in \mathbb{R}^{|\mathbb{C}|}$, where $\mathbb{C}$ is the {entity label set}.

For source language data $D_{src}$, the ground truth label $y_i$ for each span $s_i$ is available. The loss function is defined as:
\begin{equation}
    L_{src} = \frac{1}{N}\sum_{X\in D_{src}}\frac{1}{|S(X)|}\sum_{s_i\in S(X)}\text{CE}(p_i,y_i)  \label{eq:loss_src}
\end{equation}
where $\text{CE}(\cdot,\cdot)$ denotes the cross-entropy loss, and $N$ is the number of sentences in the dataset.

\subsubsection{{Masked Language Model Task}}
Due to the scarcity of labeled data in the target language, many prior studies further align the source and target languages to improve the model's performance.
{However, their emphasis remains on language-agnostic features.}
Instead, we conduct the masked language model task on target language data. As a widely used pre-trained task, it facilitates us to fully exploit the knowledge in the multilingual pre-trained language model {and improve the model's generalization ability with target language-specific features}. 

Formally, given a sentence $X=\{x_i\}_{i=1}^n$ in target language data $D_{tgt}$, we obtain the masked sentence $X^{'}=\{x_i^{'}\}_{i=1}^n$, where each token $x_i^{'}$ may be equal to either the original value $x_i$ or a special token ``[MASK]" based on a sample rate $r_m$. Then we obtain token representations $H^{'}=\{h^{'}_i\}_{i=1}^n$ for the masked sentence through the PLM. Finally, $H^{'}$ is passed through a masked language model head (MLM Head) to yield their probability distributions $p_i^{'} \in \mathbb{R}^{|\mathbb{V}|}$, where $\mathbb{V}$ is the vocabulary set of PLM.

For $x_i^{'} \in X^{'}$, its label $y_i^{'}$ is defined as the token id of $x_i \in X$ in the vocabulary set $\mathbb{V}$. The loss function is:
\begin{equation}
    \resizebox{.91\linewidth}{!}{$
            \displaystyle
                L_{mlm} = \dfrac{1}{N}\sum_{X\in D_{tgt}}\frac{1}{|X^{'}|}\sum_{x_i^{'}\in X^{'}}\mathbb{I}(x_i^{'}=[MASK])\cdot\text{CE}(p_i^{'},y_i^{'})
        $}
    \label{eq:loss_mlm}
\end{equation}
where $\mathbb{I}(\cdot)$ is the indicator function.

\subsubsection{{Training Objectives}}
In our framework, 
we simultaneously conduct the NER task on source language data and the MLM task on target language data to obtain a source model. Then, 
we use the source model to assign initial pseudo labels to the target language data. 
This enables the generation of higher-quality initial pseudo labels, establishing a robust foundation for subsequent training. Here the training objective for the source model is defined as:
\begin{equation}
    L_S = L_{src} + \omega L_{mlm} \label{eq:loss_weak}
\end{equation}
where $\omega$ is a weight hyperparameter.

\begin{figure*}[!t]
	\centering
	\includegraphics[width=1\linewidth]{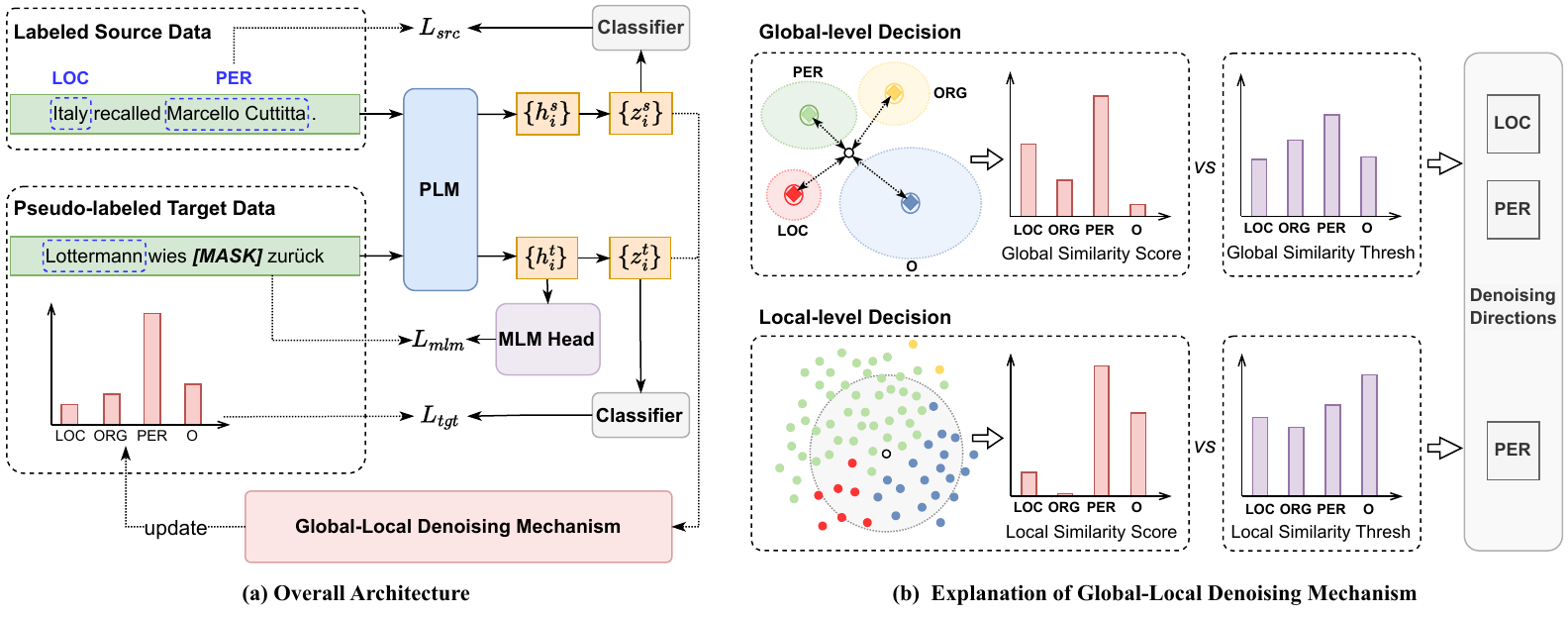}
	\caption{{(a) Overall architecture of GLoDe. The source and target language sentences are first fed to the PLM to obtain their token representations $h_i^s, h_i^t$ and following span representations $z_i^s, z_i^t$. Then $z_i^s$ and $z_i^t$ are input into a classifier for the NER task. $h_i^t$ are fed to a masked language model head (MLM Head) for the masked language model task. $z_i^s$ and $z_i^t$ are further utilized for pseudo label denoising. (b) Explanation of the global-local denoising mechanism. We first compute the similarity score of the span by its cosine distances to prototypes (global level) and its neighbors' types (local level). Then we compare the similarities with class-specific thresholds to obtain denoising directions. Decisions from both global and local levels are further integrated.}}
	\label{fig:model_main}
\end{figure*}

After obtaining the source model, we train the target model with our proposed framework. The loss function for the NER task on target language data is defined as follows:
\begin{equation}
    L_{tgt} = \frac{1}{N}\sum_{X\in D_{tgt}}\frac{1}{|S(X)|}\sum_{s_i\in S(X)}\text{CE}(p_i,\hat{p_i})  \label{eq:loss_tgt}
\end{equation}
where $\hat{p_i}$ is the soft pseudo label generated by the source model.
As shown in Figure~\ref{fig:model_main}(a), the soft pseudo label is refined by the global-local denoising mechanism. We will describe the denoising mechanism in the next section. Finally, the total objective for the target model is defined as follows:
\begin{equation}
    L_T = L_{src} + L_{tgt} + \omega L_{mlm} \label{eq:loss_all}
\end{equation}
where $\omega$ takes the same value adopted in Eq.\ref{eq:loss_weak}.

\subsection{Global-Local Denoising Mechanism}

As shown in Figure~\ref{fig:model_main}(b), two parallel decision strategies are employed to determine the final denoising directions for pseudo labels of target language data.

\subsubsection{Global-level Decision}
This module establishes a set of denoising directions based on the global information within the semantic space. The procedure is shown in the upper part of Figure~\ref{fig:model_main}(b).

\paragraph{Global Similarity Score}
We employ the prototype~\cite{snell2017prototypical} of each entity type to describe the global distribution of the semantic space. The prototype $\phi_{c}$ of entity type $c \in \mathbb{C}$ serves as the clustering center of samples belonging to this type, which is robust to outliers to some extent~\cite{xie2018learning}. Following previous works~\cite{ge2023prokd,zhou2023conproto}, we adopt a moving average scheme based on batch data to compute $\phi_{c}$, ensuring computational efficiency and a stable training process. Specially, for a batch of source and target language span representations $\{z_i\}$, $\phi_{c}$ is computed and updated as follows:
\begin{equation}
    \phi_c^{t+1} = \text{Normalize}(\alpha \phi_c^t + (1-\alpha)z_i),~\forall i \in \{i|y_i=c\}
    \label{eq:prototypes}
\end{equation}
where $t$ denotes the training step, and $y_i$ is either the ground truth label for the source language span or the pseudo label for the target language span. $\alpha$ is the update rate coefficient. {$L2$ normalization is adopted here.}

Based on these prototypes, we use the dot product to compute the similarity between a target language span $z_i$ and an entity type $c$: $sim_i^{c} = z_i\cdot \phi_c$.
Similarly, we obtain the similarities between $z_i$ and all types: $Sim_i = \{sim_i^c\}_{c \in \mathbb{C}}$, which we call ``global similarity score''.

\paragraph{Global Similarity Threshold}
We determine thresholds for each entity type to assess the aforementioned similarity scores. Instead of employing fixed thresholds, we adopt the average similarity score of each type as dynamic ones and update them at the end of each training epoch to ensure their validity and training stability. Specifically, we first segment the target language spans based on their pseudo-labeled entity types: $Z_c = \{z_i|\hat{y_i}=c, \hat{y_i}=\text{argmax}(\hat{p_i})\}$, where $\hat{p_i}$ is the soft pseudo label for $z_i$. Then the similarity threshold for type c is defined as follows:
\begin{equation}
    sim_{T}^{c} = \frac{1}{|Z_c|}\sum_{z_i \in Z_c}sim_i^c \label{eq:global_thresh}
\end{equation}
We obtain the similarity thresholds for all entity types: $Sim_{T} = \{sim_{T}^c\}_{c \in \mathbb{C}}$, called ``global similarity thresh''.

\paragraph{Global Denoising Directions}
Conventional approaches typically consider solely the type with the highest similarity score. However, a significant portion of misclassified samples are most similar to their incorrect types’ prototypes. Thus, we consider more types rather than the most similar ones. For a target language span $z_i$, we compare its similarity score $Sim_i$ with the similarity thresh $Sim_T$ to determine the update directions for its pseudo label. We describe the update directions as a binary vector: 
\begin{equation}
    d_i = [b_i^1,b_i^2,\dots,b_i^{|\mathbb{C}|}],~\text{where}~b_i^c = \mathbb{I}(sim_i^c > sim_{T}^c) 
    \label{eq:denoising_directions}
\end{equation}

Take the upper part of Figure~\ref{fig:model_main}(b) for example. Suppose we assume the entity types are \{LOC, ORG, PER, O\} (``O'' denotes non-entity), the pseudo label denoising directions (LOC and PER) can be expressed as the vector $[1,0,1,0]$.

In addition, considering the class imbalance problem (i.e., non-entity spans constitute a substantial portion)~\cite{zhou2023conproto}, we handle the non-entity type differently. If a span has the highest similarity score with type ``O'', we directly take ``O'' as an update direction for its pseudo label, regardless of whether this similarity score surpasses the threshold.

\subsubsection{Local-level Decision}
As shown in the lower part of Figure~\ref{fig:model_main}(b), the procedure of this module aligns with that of Global-level Decision. The main difference lies in the information employed.

\paragraph{Local Similarity Score}
We leverage the types of a sample's neighbors to describe the local distribution of the semantic space. 
The use of robust learning techniques, which leverage neighbor information, is extensively employed to effectively handle noisy data ~\cite{ma-etal-2023-colada}. 
We utilize the entity type distribution of a sample’s neighbors to indicate the sample’s similarity with each entity type. Specifically, we first collect source and target language span representations to construct a representation repository $D_z$. We update $D_z$ at the end of each training epoch to accommodate changes in representation distribution during the training process. Then, for a target language span representation $z_i$, we retrieve its $K$-nearest neighbors $N_k(z_i)$ from $D_z$. Finally we compute the similarity between $z_i$ and entity type $c$ based on entity types of spans in $N_k(z_i)$:
\begin{equation}
    sim_i^{c} = \frac{1}{K}\sum_{z_j \in N_k(z_i)}\mathbb{I}(y_j=c) \label{eq:local_sim}
\end{equation}
where $y_j$ is either the ground truth label for the source language span or the pseudo label for the target language span. Similarly, we obtain the similarities between $z_i$ and all types: $Sim_i = \{sim_i^c\}_{c \in \mathbb{C}}$, which we call ``local similarity score''.

\paragraph{Local Similarity Threshold and Denoising Directions}
Based on the local similarity score acquired above, we employ the same strategy and computation process for the global similarity threshold in Eq.\ref{eq:global_thresh} to obtain the local similarity threshold. Subsequently, following Eq.\ref{eq:denoising_directions}, we obtain denoising directions by utilizing the local similarity score and the local similarity threshold. We omit the detailed description here for brevity.

\subsubsection{Pseudo Label Refinement}
Noise samples exhibit a range of styles, including isolated points situated away from centroids and fuzzy points near classification boundaries. To derive the final denoising verdict, we integrate decisions made on both global and local levels. This strategy assists in detecting and addressing a variety of noise patterns.
Specifically, for a target language span $z_i$, assuming its global-level and local-level denoising direction vectors are $d_i^g$ and $d_i^l$, the update decision for its pseudo label is defined as follows: 
\begin{equation}
    u_i = \text{Normalize}(d_i^g + d_i^l) \label{eq:update_rate}
\end{equation}

Let the current pseudo label of $z_i$ be $p_i^t$. The updated pseudo label is computed as follows:
\begin{equation}
    p_i^{t+1} = \text{Normalize}(\beta p_i^t + (1-\beta)u_i) \label{eq:update_pseudo}
\end{equation}
where $\beta$ is the update rate coefficient. {$L1$ normalization is adopted in Eq.\ref{eq:update_rate} and Eq.\ref{eq:update_pseudo}}. With the improvement of pseudo labels, the referential value of the update decision $u_i$ increases. Thus, we uniformly decrease $\beta$ to increase the impact of $u_i$ during training.


There might exist multiple update directions with different updating degrees as shown in Figure~\ref{fig:model_main}(b) (e.g., PER has a higher updating degree than LOC). This allows misclassified spans to escape from their original incorrect types and adjust to high-confidence types, which are more likely to be correct. During the updating procedure, information within the semantic space is gradually integrated into the soft pseudo labels, which is beneficial for improving the model’s generalization ability to the target language data.

\section{Experiments}

\subsection{Experiment Settings}

\subsubsection{Datasets}

We conduct experiments on two widely-used benchmarks: CoNLL~\cite{sang2002conll,sang2003conll} and WikiAnn~\cite{pan2017wikiann}. The CoNLL dataset comprises four languages: English (en), German (de), Spanish (es), and Dutch (nl). The dataset is annotated with four entity types: PER, LOC, ORG, and MISC. The WikiAnn dataset comprises four languages as well: English (en), Arabic (ar), Hindi (hi), and Chinese (zh). The dataset is annotated with three entity types: PER, LOC, and ORG.

Following previous works~\cite{wu2021unitrans,zhao2022transadv}, we take the English data in each dataset as the source language data and other languages as the target language. We remove entity labels from training sets of target languages, taking them as the unlabeled target language data.

\subsubsection{Implementation Details}

Following previous works~\cite{zeng2022dualner,ma-etal-2023-colada}, we use XLM-R-large~\cite{conneau2020unsupervised} as the backbone pre-trained language model. 
The model is trained for 8 epochs (except 6 epochs for zh) and the batch size is 32. We use AdamW as the optimizer and set the learning rate to 1e-5. We set the mask rate $r_m$ to 0.25 and loss weight $\omega$ to $1e-3$ ($1e-5$ for es) for the MLM task. To improve the training efficiency, we map span representations to a low-dimensional space of 128 dimensions for the denoising computations. For neighbor retrieval, we set $K$ to 300 and utilize the FAISS tool~\cite{johnson2019billion}. The update rate coefficient $\alpha=0.99$ in Eq.\ref{eq:prototypes}. $\beta$ in Eq.\ref{eq:update_pseudo} uniformly decreases from $0.95$ to $0.80$ during training.

\subsubsection{Baselines}

We compare our method with previous baselines as follows: (1) \textbf{AdvPicker}~\cite{chen2021advpicker} employs an adversarial discriminator to select pseudo-labeled target language data with high confidence for training. (2) \textbf{RIKD}~\cite{liang2021rikd} leverages reinforcement learning for selecting pseudo-labeled data and iteratively conducts training for multiple rounds. (3) \textbf{MTMT}~\cite{li2022unsupervised} introduces an entity similarity evaluator to support the training of the NER task. (4) \textbf{CROP}~\cite{yang2022crop} transposes the NER task from the target language to the source language via back-translation and entity matching. (5) \textbf{TransAdv}~\cite{zhao2022transadv} introduces multiple data sources to obtain distinct source models for comprehensive pseudo labels. (6) \textbf{ProKD}~\cite{ge2023prokd} proposes a contrastive learning-based prototype alignment approach to boost the model’s ability to capture language-agnostic knowledge. (7) \textbf{CoLaDa}~\cite{ma-etal-2023-colada} utilizes two distinct models to denoise the labels of each other and further considers the label consistency to re-weight the noisy labels. (8) \textbf{ContProto}~\cite{zhou2023conproto} incorporates contrastive learning and prototype-based pseudo-labeling for representation learning and pseudo-label refinement.


\begin{table}[!t]
\centering
{\scriptsize
\resizebox{0.96\linewidth}{!}{
\begin{tabular}{@{}lllll@{}}
\toprule
\textbf{Method}     & \multicolumn{1}{c}{\textbf{de}} & \multicolumn{1}{c}{\textbf{es}} & \multicolumn{1}{c}{\textbf{nl}} & \multicolumn{1}{c}{\textbf{Avg}}   \\ \midrule
AdvPicker           & 75.01          & 79.00          & 82.90          & 78.97          \\
RIKD                & 75.48          & 77.84          & 82.46          & 78.59          \\
MTMT                & 76.80          & 81.82          & 83.41          & 80.68          \\
CROP                & 80.10          & 78.10          & 79.50          & 79.23          \\
TransAdv            & 75.52          & 80.93          & 83.78          & 80.08          \\
ProKD               & 78.90          & 82.62          & 79.53          & 80.35          \\
CoLaDa              & \textbf{81.12} & 82.70          & 85.15          & 82.99          \\
ContProto           & 76.41          & 85.02          & 83.69          & 81.71          \\ \midrule
\textbf{GLoDe(Ours)} & 79.15          & \textbf{85.29} & \textbf{85.76} & \textbf{83.40} \\ \bottomrule
\end{tabular}
}
}
\caption{{Experimental results on CoNLL. Results on German (de), Spanish (es), Dutch(nl), and their average performance are reported.}}
\label{tab:conll-res}
\end{table}

\begin{table}[!t]
\centering
{\scriptsize
\resizebox{0.96\linewidth}{!}{
\begin{tabular}{@{}lllll@{}}
\toprule
\textbf{Method}     & \multicolumn{1}{c}{\textbf{ar}} & \multicolumn{1}{c}{\textbf{hi}} & \multicolumn{1}{c}{\textbf{zh}} & \multicolumn{1}{c}{\textbf{Avg}}   \\ \midrule
AdvPicker$^\dagger$           & 39.75          & 71.94          & 55.24          & 55.65          \\
RIKD                & 45.96          & 70.28          & 50.40          & 55.55          \\
MTMT                & 52.77          & 70.76          & 52.26          & 58.60          \\
CROP                & 48.00          & 77.10          & 50.30          & 58.47          \\
TransAdv            & 42.53          & 74.24          & 54.25          & 49.17          \\
ProKD               & 50.91          & 70.72          & 51.80          & 57.81          \\
CoLaDa              & 66.94          & 76.69          & 60.08          & 67.90          \\
ContProto           & 72.20          & 83.45          & 61.47          & 72.37          \\ \midrule
\textbf{GLoDe(Ours)} & \textbf{74.35} & \textbf{83.61} & \textbf{64.81} & \textbf{74.26} \\ \bottomrule
\end{tabular}
}
}
\caption{{Experimental results on WikiAnn. Results on Arabic (ar), Hindi (hi), Chinese (zh), and their average performance are reported. $\dagger$ denotes results obtained by running their public code on the data.}}
\label{tab:wikiann-res}
\end{table}

\subsection{Main Results}

Tables~\ref{tab:conll-res} and~\ref{tab:wikiann-res} present the results of our proposed method and previous baselines on CoNLL and WikiAnn. Following previous works, we employ the entity-level micro-F1 score as the metric. We conducted all experiments three times and calculated the average value as the final result.

\begin{table*}[h]
\centering
\resizebox{0.98\linewidth}{!}{
\begin{tabular}{@{}lllllll@{}}
\toprule
\textbf{Method}             & \textbf{de}    & \textbf{es}    & \textbf{nl}    & \textbf{ar}    & \textbf{hi}    & \textbf{zh}    \\ \midrule
\textbf{GLoDe}               & \textbf{79.15} & \textbf{85.29} & \textbf{85.76} & \textbf{74.35} & \textbf{83.61} & \textbf{64.81} \\ \midrule
- w/o mlm                   & 77.48(-1.67)   & 84.28(-1.01)   & 84.63(-1.13)   & 72.26(-2.09)   & 82.08(-1.53)   & 63.47(-1.34)   \\ \midrule
- w/o denoising             & 77.07(-2.08)   & 84.11(-1.18)   & 84.04(-1.72)   & 70.83(-3.52)   & 82.64(-0.97)   & 63.44(-1.37)   \\
- w/o denoising-local      & 77.49(-1.66)   & 84.26(-1.03)   & 84.46(-1.30)   & 71.85(-2.50)   & 81.69(-1.92)   & 64.03(-0.78)   \\
- w/o denoising-global     & 77.81(-1.34)   & 84.18(-1.11)   & 84.82(-0.94)   & 71.86(-2.49)   & 82.42(-1.19)   & 64.37(-0.44)   \\
- w/o multiple directions & 77.41(-1.74)   & 84.20(-1.09)   & 84.23(-1.53)   & 71.24(-3.11)   & 82.50(-1.11)   & 63.46(-1.35)   \\ \midrule
- w/o mlm init               & 74.75(-4.40)   & 84.05(-1.25)   & 84.76(-1.00)   & 65.35(-9.00)   & 81.29(-2.32)   & 61.07(-3.74)   \\ \bottomrule
\end{tabular}
}
\caption{{Ablation studies on CoNLL and WikiAnn dataset. In this table, we remove specific components in our model for comparison. Values in brackets denote the performance drop compared to the full model.}}
\label{tab:ablation}
\end{table*}

It can be seen that our proposed GLoDe outperforms prior methods, achieving an improvement of 0.41 F1 scores on average for CoNLL and 1.89 F1 scores on average for WikiAnn. This demonstrates the effectiveness of our approach. Besides, 
there are two noteworthy findings evident in the results: (1) Our model reaches suboptimal performance on German (de). The German dataset has fewer entities, and numerous sentences do not contain entities, potentially leading the model to predict more false positive entities. Additionally, both CoLaDa and CROP incorporate translated content, which facilitates the model to acquire more entity information in the target language. (2) For target languages (e.g., es, nl) closer to the source language (en), previous works have shown incremental improvements in a moderate trend. Conversely, for distant source-target pairs (e.g., en-ar, en-zh), models performing the denoising process on pseudo-labeled target language data (ours, ContProto, and CoLaDa) have shown significant improvements over other models.
We attribute this to the critical label noise problem in these distant languages.


\subsection{Ablation Study}

To further validate the effectiveness of components in our proposed model, we conduct the following ablation studies: (1) \textit{w/o mlm}, which removes the masked language model task during training; (2) \textit{w/o denoising}, which removes the global-local denoising mechanism; (3) \textit{w/o denoising-local}, which only leverages global information in the semantic space for pseudo label denoising; (4) \textit{w/o denoising-global}, which only leverages local information in the semantic space for pseudo label denoising; (5) \textit{w/o multiple directions}, which only considers the type with the highest similarity score as the potential update direction for the pseudo label; (6) \textit{w/o mlm init}, which omits the masked language model task, training the source model solely on source language data to acquire the initial pseudo labels.

Table~\ref{tab:ablation} shows the ablation study results. It can be seen that: 
(1) Both \textit{w/o mlm} and \textit{w/o denoising} diminish the model's performance, which demonstrates the effectiveness of these two modules. Both of them improve the model's generalization ability on the target language. Additionally, removing the denoising mechanism exerts a more significant impact on most datasets, emphasizing the significance of noise reduction for pseudo-labeled data.
(2) \textit{w/o denoising-local} and \textit{w/o denoising-global} bring less performance decline compared to \textit{w/o denoising}. This implies that leveraging either local or global information in the semantic space aids in pseudo-label denoising. Naturally, employing both together yields superior results. Moreover, there seems to be no absolute answer to the question regarding which part of the information is more useful. (3) \textit{w/o multiple directions} also improves the performance compared to \textit{w/o denoising},  but it falls short of comparison to other denoising schemes. This implies that incorporating multiple denoising directions would be more beneficial.
(4) \textit{w/o mlm init} denotes the significance of improving the source model with target language-specific features. Additionally, the quality of the initial pseudo labels significantly constrains the target model's performance.

\begin{figure}[!t]
	\centering
	\includegraphics[width=0.99\linewidth]{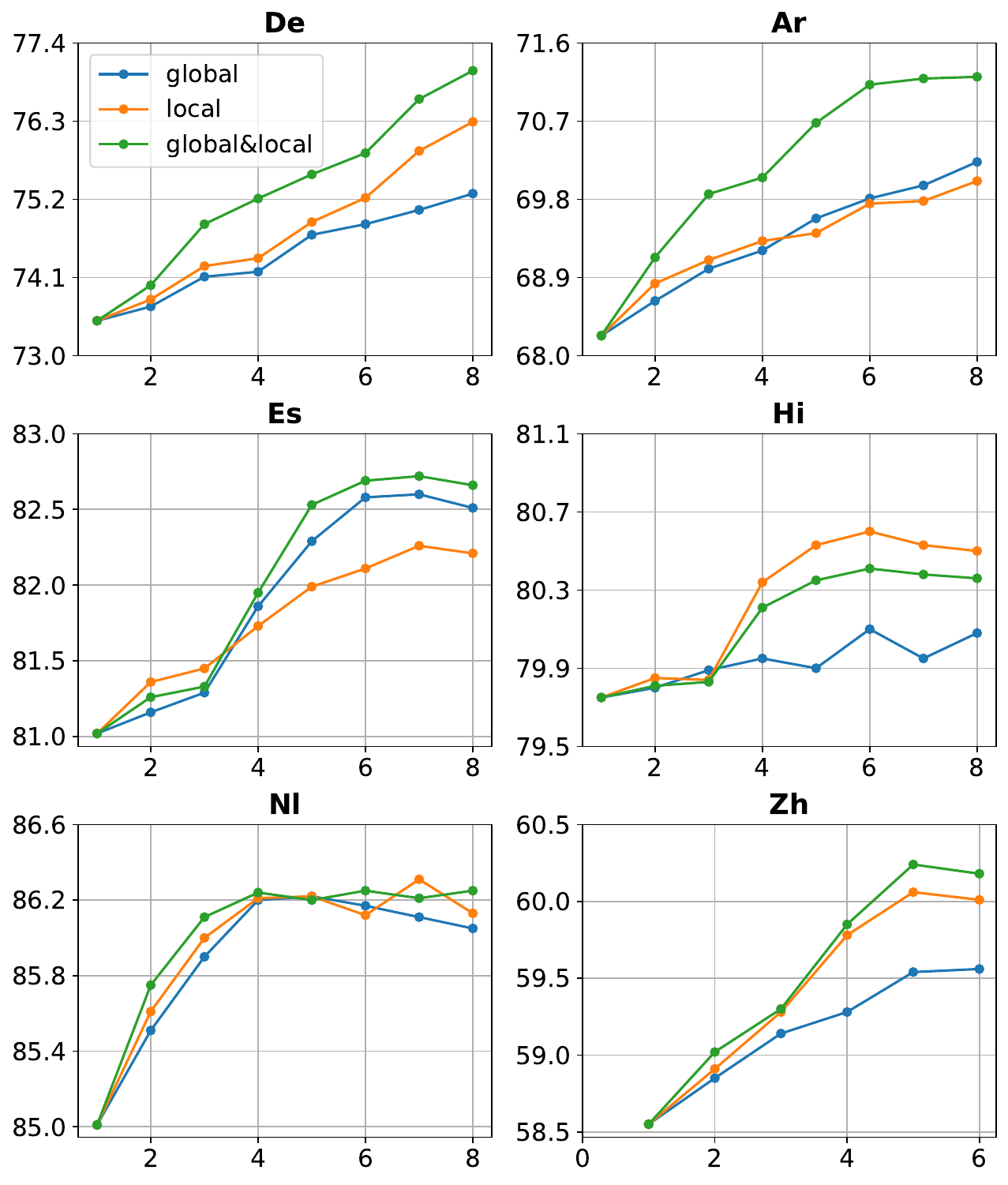}
	\caption{Quality of pseudo labels. The horizontal axis is the epoch number and the vertical axis is the entity-level F1 scores of pseudo labels. Three types of information for denoising are compared.}
	\label{fig:quality-pseu}
\end{figure}

\subsection{Quality of Pseudo Labels}

In this section, we assess the quality of pseudo labels during training. As in the prior experiments, we measure the quality by entity-level micro-F1 scores. 
We compare the following three pseudo label denoising strategies: (1)~\textit{global}, which only leverages global information in the semantic space for pseudo label denoising; (2)~\textit{local}, which only leverages local information; (3)~\textit{global\&local}, which leverages both global and local information, serving as the solution we adopt.

{As shown in Figure~\ref{fig:quality-pseu}, the quality of pseudo labels improves progressively during training. The utilization of either global or local information improves pseudo labels, with the combined use of both resulting in even greater improvement for most of the target languages. We argue that the two types of information can complement each other and they have different impacts for different target languages. For example, \textit{local} yields more advantages for De and Zh, while \textit{global} provides greater benefits for Es. For Nl, little difference is observed among the three strategies, with limited improvement obtained after 4th epoch. We attribute this to the high similarity between Nl and the source language (En). The information from either the global level or the local level may be sufficient. For Hi, \textit{global} may even exhibit a negative impact. Given that the source language contains nearly five times as many entities as Hi, local-level information proves to be more helpful than global-level information.}

\newif\ifauxiliary
\auxiliaryfalse

\ifauxiliary
\subsection{Effectiveness of Auxiliary Task}

The auxiliary task improves both the source model and the target model with target language-specific features. In this section, we analyze the impact of the auxiliary task by comparing entity distributions from different data sources. 
We take De and Ar as examples, as shown in Figure~\ref{fig:label-dstr}.

Specifically, we compare the source language entity distributions (blue lines in figures), target language entity distributions (green lines), and model-predicted entity distributions for the target language (orange lines). Predicted distributions in Figure~\ref{fig:label-dstr}(a) and (c) are obtained from models trained solely with source language data (\textbf{plain}-model). Predicted distributions in Figure~\ref{fig:label-dstr}(b) and (d) are obtained from models trained with both source language data and the auxiliary task (\textbf{aux}-model). For De, the predicted distribution of the plain model closely aligns with the source language. After introducing the auxiliary task, it shifts to the target language. For Ar, the prediction of the plain model is even dissimilar to the source language, due to the substantial gap between the source and target language. After introducing the auxiliary task, it shifts to the target language as well. These results demonstrate the effectiveness of the auxiliary task, enabling the model to capture more information about the target language distribution.

\begin{figure}[!t]
	\centering
	\includegraphics[width=0.99\linewidth]{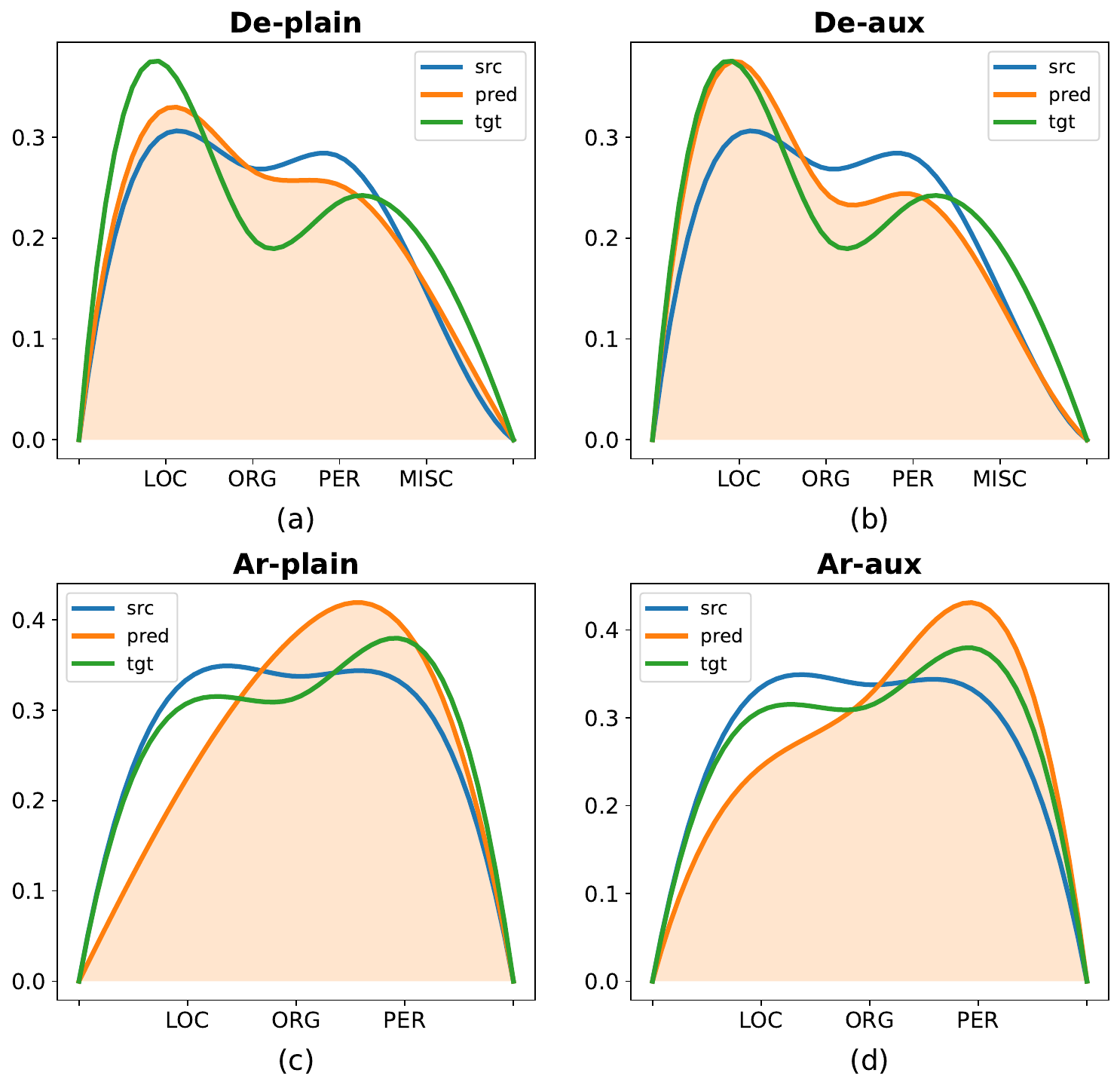}
	\caption{Distribution of entity types. The horizontal axis is the entity type and the vertical axis is the percentage of the type. \textcolor{blue}{We interpolate discrete statistical results and represent them with continuous curves for better visualization.}}
	\label{fig:label-dstr}
\end{figure}
\fi

\newif\ifagnostic
\agnosticfalse
\agnostictrue

\ifagnostic
\subsection{Effectiveness of Language-specific Features}

The auxiliary task improves both the source and target model with target language-specific features. In this section, we further analyze this impact.

As shown in Figure~\ref{fig:label-dstr-single}, we compare entity distributions from different data sources. Specifically, the source language entity distributions (blue lines), target language entity distributions (green lines), and model-predicted entity distributions for the target language (orange lines) are considered. The predicted distribution in Figure~\ref{fig:label-dstr-single}(a) is obtained from the model trained solely with source language data (\textbf{Plain}-model), and that in Figure~\ref{fig:label-dstr-single}(b) is obtained from the model trained with both source language data and the auxiliary task (\textbf{Aux}-model). We can observe that the predicted distribution shifts towards the target language distribution after introducing target language-specific features by the auxiliary task.

Additionally, we conducted some comparative experiments to demonstrate the advantages of language-specific features. We replace the masked language model task in our framework with the supervised contrastive learning task, which has recently been widely used to improve the model with language-agnostic features (\cite{ge2023prokd,zhou2023conproto}). As shown in Table~\ref{tab:target-agnstic-1}, we denote the model improved by language-agnostic features and language-specific features as $\text{GLoDe}_{\text{agnostic}}$ and $\text{GLoDe}_{\text{specific}}$, respectively. Comparison results demonstrate that language-specific features are more beneficial for model improvement.

\begin{figure}[!t]
    \centering
    \includegraphics[width=0.99\linewidth]{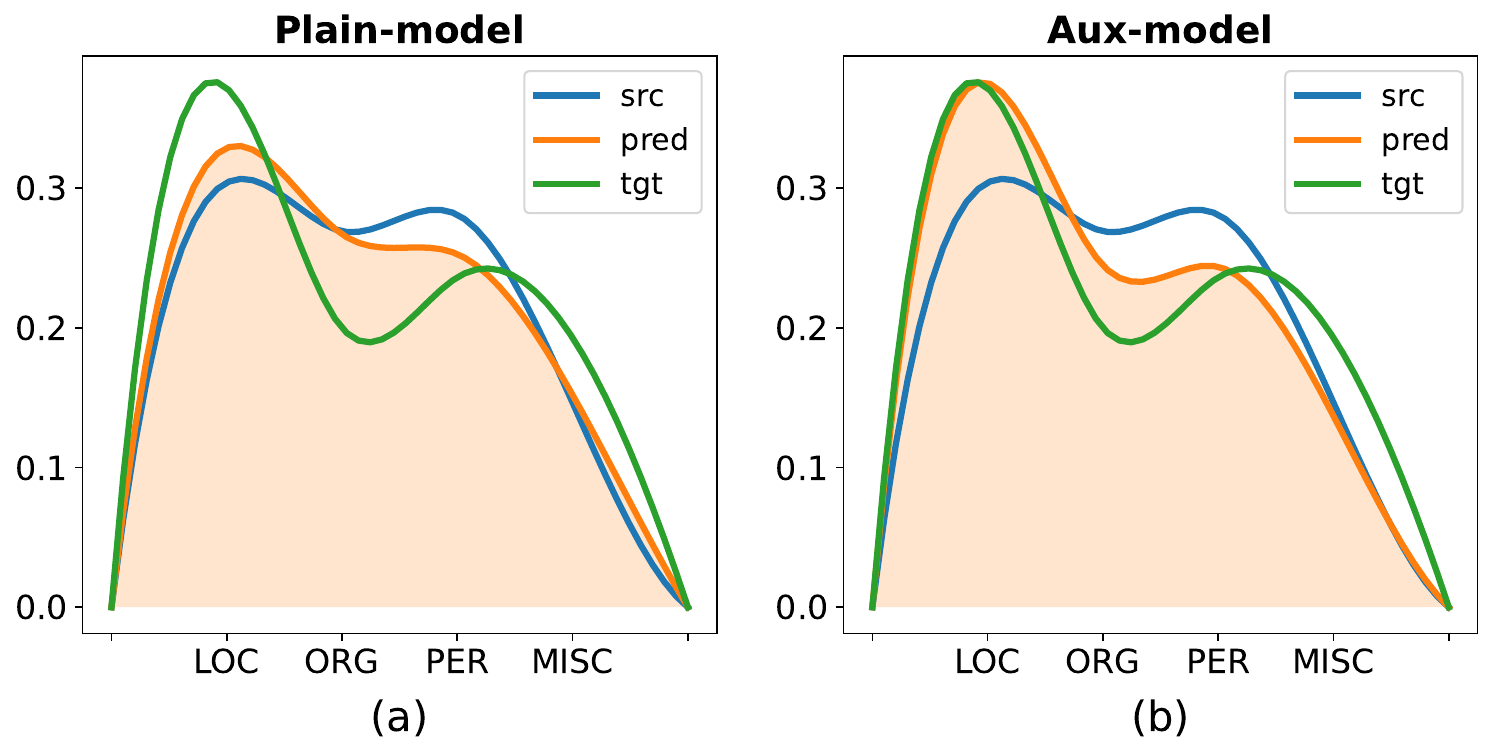}
    \caption{Distribution of entity types. The vertical axis is the percentage of the entity type. We take De as the target language here. We interpolate discrete statistical results and represent them with continuous curves for better visualization.}
    \label{fig:label-dstr-single}
\end{figure}
\begin{table}[!t]
\centering
\resizebox{0.99\linewidth}{!}{
\begin{tabular}{@{}lllllll@{}}
\toprule
\textbf{Method}          & \multicolumn{1}{c}{\textbf{de}} & \multicolumn{1}{c}{\textbf{es}} & \multicolumn{1}{c}{\textbf{nl}} & \multicolumn{1}{c}{\textbf{ar}} & \multicolumn{1}{c}{\textbf{hi}} & \multicolumn{1}{c}{\textbf{zh}} \\ \midrule
$\text{GLoDe}_{\text{specific}}$    & 79.15    & 85.29    & 85.76    & 74.35    & 83.61    & 64.81 \\
$\text{GLoDe}_{\text{agnostic}}$    & 77.52    & 84.03    & 83.81    & 72.31    & 82.71    & 62.76 \\ \bottomrule
\end{tabular}
}
\caption{Comparision of impacts of target language-specific features and language-agnostic features for model improvement.}
\label{tab:target-agnstic-1}
\end{table}

\fi

\section{Conclusion}

In this paper, we present GLoDe, a pseudo-label denoising framework for cross-lingual NER. 
Our global-local denoising mechanism integrates global and local information within the semantic space to rectify as many incorrect pseudo labels for target language data as possible.
The information within pseudo labels progressively increases during the denoising process, which helps to improve the model’s generalization ability on the target language. 
Additionally, we further improve our model with target language-specific features, leveraging the masked language model task.
Experimental results on six benchmark datasets demonstrate that GLoDe achieves superior performance over previous baselines.

\section*{Acknowledgements}
This work was supported in part by the National Natural Science Foundation of China under Grant No. 62276110, No. 62172039 and in part by the fund of Joint Laboratory of HUST and Pingan Property \& Casualty Research (HPL). The authors would also like to thank the anonymous reviewers for their comments on improving the quality of this paper. 



\bibliographystyle{named}
\bibliography{ijcai24}

\begin{thebibliography}{}

\bibitem[\protect\citeauthoryear{Chen \bgroup \em et al.\egroup }{2021}]{chen2021advpicker}
Weile Chen, Huiqiang Jiang, Qianhui Wu, B{\"o}rje Karlsson, and Yi~Guan.
\newblock Advpicker: Effectively leveraging unlabeled data via adversarial discriminator for cross-lingual ner.
\newblock In {\em Proceedings of the 59th Annual Meeting of the Association for Computational Linguistics and the 11th International Joint Conference on Natural Language Processing}, pages 743--753, 2021.

\bibitem[\protect\citeauthoryear{Conneau \bgroup \em et al.\egroup }{2020}]{conneau2020unsupervised}
Alexis Conneau, Kartikay Khandelwal, Naman Goyal, Vishrav Chaudhary, Guillaume Wenzek, Francisco Guzm{\'a}n, {\'E}douard Grave, Myle Ott, Luke Zettlemoyer, and Veselin Stoyanov.
\newblock Unsupervised cross-lingual representation learning at scale.
\newblock In {\em Proceedings of the 58th Annual Meeting of the Association for Computational Linguistics}, pages 8440--8451, 2020.

\bibitem[\protect\citeauthoryear{Fan \bgroup \em et al.\egroup }{2023}]{fan2023improving}
Chenghao Fan, Wei Wei, Xiaoye Qu, Zhenyi Lu, Wenfeng Xie, Yu~Cheng, and Dangyang Chen.
\newblock Improving low-resource prompt-based relation representation with multi-view decoupling learning.
\newblock {\em arXiv preprint arXiv:2312.17267}, 2023.

\bibitem[\protect\citeauthoryear{Fu \bgroup \em et al.\egroup }{2021}]{fu2021spanner}
Jinlan Fu, Xuan-Jing Huang, and Pengfei Liu.
\newblock Spanner: Named entity re-/recognition as span prediction.
\newblock In {\em Proceedings of the 59th Annual Meeting of the Association for Computational Linguistics and the 11th International Joint Conference on Natural Language Processing}, pages 7183--7195, 2021.

\bibitem[\protect\citeauthoryear{Ge \bgroup \em et al.\egroup }{2023}]{ge2023prokd}
Ling Ge, Chunming Hu, Guanghui Ma, Hong Zhang, and Jihong Liu.
\newblock Prokd: an unsupervised prototypical knowledge distillation network for zero-resource cross-lingual named entity recognition.
\newblock In {\em Proceedings of the AAAI Conference on Artificial Intelligence}, volume~37, pages 12818--12826, 2023.

\bibitem[\protect\citeauthoryear{Gu \bgroup \em et al.\egroup }{2022}]{gu2022delving}
Yingjie Gu, Xiaoye Qu, Zhefeng Wang, Yi~Zheng, Baoxing Huai, and Nicholas~Jing Yuan.
\newblock Delving deep into regularity: A simple but effective method for chinese named entity recognition.
\newblock In {\em Findings of the Association for Computational Linguistics: NAACL 2022}, pages 1863--1873, 2022.

\bibitem[\protect\citeauthoryear{Hinton \bgroup \em et al.\egroup }{2015}]{hinton2015distilling}
Geoffrey Hinton, Oriol Vinyals, and Jeff Dean.
\newblock Distilling the knowledge in a neural network.
\newblock {\em arXiv preprint arXiv:1503.02531}, 2015.

\bibitem[\protect\citeauthoryear{Jain \bgroup \em et al.\egroup }{2019}]{jain2019entity}
Alankar Jain, Bhargavi Paranjape, and Zachary~C Lipton.
\newblock Entity projection via machine translation for cross-lingual ner.
\newblock In {\em Proceedings of the 2019 Conference on Empirical Methods in Natural Language Processing and the 9th International Joint Conference on Natural Language Processing}, pages 1083--1092, 2019.

\bibitem[\protect\citeauthoryear{Johnson \bgroup \em et al.\egroup }{2019}]{johnson2019billion}
Jeff Johnson, Matthijs Douze, and Herv{\'e} J{\'e}gou.
\newblock Billion-scale similarity search with gpus.
\newblock {\em IEEE Transactions on Big Data}, 7(3):535--547, 2019.

\bibitem[\protect\citeauthoryear{Keung \bgroup \em et al.\egroup }{2019}]{keung2019adversarial}
Phillip Keung, Yichao Lu, and Vikas Bhardwaj.
\newblock Adversarial learning with contextual embeddings for zero-resource cross-lingual classification and ner.
\newblock In {\em Proceedings of the 2019 Conference on Empirical Methods in Natural Language Processing and the 9th International Joint Conference on Natural Language Processing (EMNLP-IJCNLP)}, pages 1355--1360, 2019.

\bibitem[\protect\citeauthoryear{Li \bgroup \em et al.\egroup }{2022}]{li2022unsupervised}
Zhuoran Li, Chunming Hu, Xiaohui Guo, Junfan Chen, Wenyi Qin, and Richong Zhang.
\newblock An unsupervised multiple-task and multiple-teacher model for cross-lingual named entity recognition.
\newblock In {\em Proceedings of the 60th Annual Meeting of the Association for Computational Linguistics}, pages 170--179, 2022.

\bibitem[\protect\citeauthoryear{Liang \bgroup \em et al.\egroup }{2021}]{liang2021rikd}
Shining Liang, Ming Gong, Jian Pei, Linjun Shou, Wanli Zuo, Xianglin Zuo, and Daxin Jiang.
\newblock Reinforced iterative knowledge distillation for cross-lingual named entity recognition.
\newblock In {\em Proceedings of the 27th ACM SIGKDD Conference on Knowledge Discovery \& Data Mining}, pages 3231--3239, 2021.

\bibitem[\protect\citeauthoryear{Liu \bgroup \em et al.\egroup }{2021}]{liu2021mulda}
Linlin Liu, Bosheng Ding, Lidong Bing, Shafiq Joty, Luo Si, and Chunyan Miao.
\newblock Mulda: A multilingual data augmentation framework for low-resource cross-lingual ner.
\newblock In {\em Proceedings of the 59th Annual Meeting of the Association for Computational Linguistics and the 11th International Joint Conference on Natural Language Processing}, pages 5834--5846, 2021.

\bibitem[\protect\citeauthoryear{Ma \bgroup \em et al.\egroup }{2022}]{ma2022wider}
Jun-Yu Ma, Beiduo Chen, Jia-Chen Gu, Zhenhua Ling, Wu~Guo, Quan Liu, Zhigang Chen, and Cong Liu.
\newblock Wider \& closer: Mixture of short-channel distillers for zero-shot cross-lingual named entity recognition.
\newblock In {\em Proceedings of the 2022 Conference on Empirical Methods in Natural Language Processing}, pages 5171--5183, 2022.

\bibitem[\protect\citeauthoryear{Ma \bgroup \em et al.\egroup }{2023}]{ma-etal-2023-colada}
Tingting Ma, Qianhui Wu, Huiqiang Jiang, B{\"o}rje Karlsson, Tiejun Zhao, and Chin-Yew Lin.
\newblock {C}o{L}a{D}a: A collaborative label denoising framework for cross-lingual named entity recognition.
\newblock In {\em Proceedings of the 61st Annual Meeting of the Association for Computational Linguistics (Volume 1: Long Papers)}, pages 5995--6009, Toronto, Canada, July 2023.

\bibitem[\protect\citeauthoryear{Mayhew \bgroup \em et al.\egroup }{2017}]{mayhew2017cheap}
Stephen Mayhew, Chen-Tse Tsai, and Dan Roth.
\newblock Cheap translation for cross-lingual named entity recognition.
\newblock In {\em Proceedings of the 2017 conference on empirical methods in natural language processing}, pages 2536--2545, 2017.

\bibitem[\protect\citeauthoryear{Pan \bgroup \em et al.\egroup }{2017}]{pan2017wikiann}
Xiaoman Pan, Boliang Zhang, Jonathan May, Joel Nothman, Kevin Knight, and Heng Ji.
\newblock Cross-lingual name tagging and linking for 282 languages.
\newblock In {\em Proceedings of the 55th Annual Meeting of the Association for Computational Linguistics}, pages 1946--1958, 2017.

\bibitem[\protect\citeauthoryear{Pires \bgroup \em et al.\egroup }{2019}]{pires2019multilingual}
Telmo Pires, Eva Schlinger, and Dan Garrette.
\newblock How multilingual is multilingual bert?
\newblock In {\em Proceedings of the 57th Annual Meeting of the Association for Computational Linguistics}, pages 4996--5001, 2019.

\bibitem[\protect\citeauthoryear{Qu \bgroup \em et al.\egroup }{2023a}]{qu2023survey}
Xiaoye Qu, Yingjie Gu, Qingrong Xia, Zechang Li, Zhefeng Wang, and Baoxing Huai.
\newblock A survey on arabic named entity recognition: Past, recent advances, and future trends.
\newblock {\em IEEE Transactions on Knowledge and Data Engineering}, 2023.

\bibitem[\protect\citeauthoryear{Qu \bgroup \em et al.\egroup }{2023b}]{qu2023distantly}
Xiaoye Qu, Jun Zeng, Daizong Liu, Zhefeng Wang, Baoxing Huai, and Pan Zhou.
\newblock Distantly-supervised named entity recognition with adaptive teacher learning and fine-grained student ensemble.
\newblock In {\em Proceedings of the AAAI Conference on Artificial Intelligence}, volume~37, pages 13501--13509, 2023.

\bibitem[\protect\citeauthoryear{Sang and De~Meulder}{2003}]{sang2003conll}
Erik Tjong~Kim Sang and Fien De~Meulder.
\newblock Introduction to the conll-2003 shared task: Language-independent named entity recognition.
\newblock In {\em Proceedings of the Seventh Conference on Natural Language Learning at HLT-NAACL 2003}, pages 142--147, 2003.

\bibitem[\protect\citeauthoryear{Sang and Erik}{2002}]{sang2002conll}
Tjong~Kim Sang and F~Erik.
\newblock Introduction to the conll-2002 shared task: language-independent named entity recognition.
\newblock In {\em Proceedings of CoNLL-2002/Roth, Dan [edit.]}, pages 155--158, 2002.

\bibitem[\protect\citeauthoryear{Snell \bgroup \em et al.\egroup }{2017}]{snell2017prototypical}
Jake Snell, Kevin Swersky, and Richard Zemel.
\newblock Prototypical networks for few-shot learning.
\newblock {\em Advances in neural information processing systems}, 2017.

\bibitem[\protect\citeauthoryear{Tsai \bgroup \em et al.\egroup }{2016}]{tsai2016wikification}
Chen-Tse Tsai, Stephen Mayhew, and Dan Roth.
\newblock Cross-lingual named entity recognition via wikification.
\newblock In {\em Proceedings of The 20th SIGNLL Conference on Computational Natural Language Learning}, pages 219--228, 2016.

\bibitem[\protect\citeauthoryear{Wei \bgroup \em et al.\egroup }{2021}]{wei2021position}
Wei Wei, Zanbo Wang, Xianling Mao, Guangyou Zhou, Pan Zhou, and Sheng Jiang.
\newblock Position-aware self-attention based neural sequence labeling.
\newblock {\em Pattern Recognition}, 110:107636, 2021.

\bibitem[\protect\citeauthoryear{Wu and Dredze}{2020}]{wu2020explicit}
Shijie Wu and Mark Dredze.
\newblock Do explicit alignments robustly improve multilingual encoders?
\newblock In {\em Proceedings of the 2020 Conference on Empirical Methods in Natural Language Processing (EMNLP)}, pages 4471--4482, 2020.

\bibitem[\protect\citeauthoryear{Wu \bgroup \em et al.\egroup }{2020}]{wu2020tsl}
Qianhui Wu, Zijia Lin, B{\"o}rje Karlsson, Jian-Guang Lou, and Biqing Huang.
\newblock Single-/multi-source cross-lingual ner via teacher-student learning on unlabeled data in target language.
\newblock In {\em Proceedings of the 58th Annual Meeting of the Association for Computational Linguistics}, pages 6505--6514, 2020.

\bibitem[\protect\citeauthoryear{Wu \bgroup \em et al.\egroup }{2021}]{wu2021unitrans}
Qianhui Wu, Zijia Lin, B{\"o}rje~F Karlsson, Biqing Huang, and Jian-Guang Lou.
\newblock Unitrans: unifying model transfer and data transfer for cross-lingual named entity recognition with unlabeled data.
\newblock In {\em Proceedings of the Twenty-Ninth International Conference on International Joint Conferences on Artificial Intelligence}, pages 3926--3932, 2021.

\bibitem[\protect\citeauthoryear{WU~SJ}{2019}]{wu2019surprising}
DREDZE~M WU~SJ.
\newblock The surprising cross-lingual effectiveness of bert.
\newblock In {\em Proceedings of the 2019 Conference on Empirical Methods in Natural Language Processing and the 9th International Joint Conference on Natural Language Processing, Hong Kong, China}, pages 833--844, 2019.

\bibitem[\protect\citeauthoryear{Xie \bgroup \em et al.\egroup }{2018a}]{xie2018neural}
Jiateng Xie, Zhilin Yang, Graham Neubig, Noah~A Smith, and Jaime~G Carbonell.
\newblock Neural cross-lingual named entity recognition with minimal resources.
\newblock In {\em Proceedings of the 2018 Conference on Empirical Methods in Natural Language Processing}, pages 369--379, 2018.

\bibitem[\protect\citeauthoryear{Xie \bgroup \em et al.\egroup }{2018b}]{xie2018learning}
Shaoan Xie, Zibin Zheng, Liang Chen, and Chuan Chen.
\newblock Learning semantic representations for unsupervised domain adaptation.
\newblock In {\em International conference on machine learning}, pages 5423--5432, 2018.

\bibitem[\protect\citeauthoryear{Yang \bgroup \em et al.\egroup }{2022}]{yang2022crop}
Jian Yang, Shaohan Huang, Shuming Ma, Yuwei Yin, Li~Dong, Dongdong Zhang, Hongcheng Guo, Zhoujun Li, and Furu Wei.
\newblock Crop: Zero-shot cross-lingual named entity recognition with multilingual labeled sequence translation.
\newblock In {\em Findings of the Association for Computational Linguistics: EMNLP 2022}, 2022.

\bibitem[\protect\citeauthoryear{Zeng \bgroup \em et al.\egroup }{2022}]{zeng2022dualner}
Jiali Zeng, Yufan Jiang, Yongjing Yin, Xu~Wang, Binghuai Lin, and Yunbo Cao.
\newblock Dualner: A dual-teaching framework for zero-shot cross-lingual named entity recognition.
\newblock In {\em Findings of the Association for Computational Linguistics: EMNLP 2022}, pages 1837--1843, 2022.

\bibitem[\protect\citeauthoryear{Zhao \bgroup \em et al.\egroup }{2022}]{zhao2022transadv}
Yichun Zhao, Jintao Du, Gongshen Liu, and Huijia Zhu.
\newblock Transadv: A translation-based adversarial learning framework for zero-resource cross-lingual named entity recognition.
\newblock In {\em Findings of the Association for Computational Linguistics: EMNLP 2022}, pages 742--749, 2022.

\bibitem[\protect\citeauthoryear{Zhou \bgroup \em et al.\egroup }{2023}]{zhou2023conproto}
Ran Zhou, Xin Li, Lidong Bing, Erik Cambria, and Chunyan Miao.
\newblock Improving self-training for cross-lingual named entity recognition with contrastive and prototype learning.
\newblock In {\em Proceedings of the 61st Annual Meeting of the Association for Computational Linguistics}, pages 4018--4031, Toronto, Canada, July 2023. Association for Computational Linguistics.

\bibitem[\protect\citeauthoryear{Zhu \bgroup \em et al.\egroup }{2023}]{zhu2023mirror}
Tong Zhu, Junfei Ren, Zijian Yu, Mengsong Wu, Guoliang Zhang, Xiaoye Qu, Wenliang Chen, Zhefeng Wang, Baoxing Huai, and Min Zhang.
\newblock Mirror: A universal framework for various information extraction tasks.
\newblock {\em arXiv e-prints}, pages arXiv--2311, 2023.

\bibitem[\protect\citeauthoryear{Zirikly and Hagiwara}{2015}]{zirikly2015cross}
Ayah Zirikly and Masato Hagiwara.
\newblock Cross-lingual transfer of named entity recognizers without parallel corpora.
\newblock In {\em Proceedings of the 53rd Annual Meeting of the Association for Computational Linguistics and the 7th International Joint Conference on Natural Language Processing}, pages 390--396, 2015.

\end{thebibliography}

\newpage
\appendix
\section{Details of Datasets}
This section details the dataset information. The number of sentences and entities for each language are presented in Table~\ref{tab:datasets}.
As we stated previously, the entity number of the German dataset is significantly fewer than that of other languages in CoNLL, and the source language (English, WikiAnn) contains nearly five times as many entities as Hindi.

\begin{table}[!h]
\centering
\resizebox{0.99\linewidth}{!}{
\begin{tabular}{@{}cccccc@{}}
\toprule
\textbf{Datasets}                 & \textbf{Language} & \textbf{Type}     & \textbf{Train}  & \textbf{Dev}    & \textbf{Test}   \\ \midrule
\multirow{8}{*}{CoNLL}   & English  & Sentence & 14,987 & 3,466  & 3,684  \\
                         & (en)     & Entity   & 23,499 & 5,942  & 5,648  \\ \cmidrule(l){2-6} 
                         & German   & Sentence & 12,705 & 3,068  & 3,160  \\
                         & (de)     & Entity   & 11,851 & 4,833  & 3,673  \\ \cmidrule(l){2-6} 
                         & Spanish  & Sentence & 8,323  & 1,915  & 1,517  \\
                         & (es)     & Entity   & 18,798 & 4,351  & 3,558  \\ \cmidrule(l){2-6} 
                         & Dutch    & Sentence & 15,806 & 2,895  & 5,195  \\
                         & (nl)     & Entity   & 13,344 & 2,616  & 3,941  \\ \midrule
\multirow{8}{*}{WikiAnn} & English  & Sentence & 20,000 & 10,000 & 10,000 \\
                         & (en)     & Entity   & 27,931 & 14,146 & 13,958 \\ \cmidrule(l){2-6} 
                         & Arabic   & Sentence & 20,000 & 10,000 & 10,000 \\
                         & (ar)     & Entity   & 22,500 & 11,266 & 11,259 \\ \cmidrule(l){2-6} 
                         & Hindi    & Sentence & 5,000  & 1,000  & 1,000  \\
                         & (hi)     & Entity   & 6,124  & 1,226  & 1,228  \\ \cmidrule(l){2-6} 
                         & Chinese  & Sentence & 20,000 & 10,000 & 10,000 \\
                         & (zh)     & Entity   & 25,031 & 12,493 & 12,532 \\ \bottomrule
\end{tabular}
}
\caption{Statistics of the CoNLL and WikiAnn datasets.}
\label{tab:datasets}
\end{table}

\section{Model-predicted Entity Distributions}
\begin{figure*}[!t]
    \centering
    \includegraphics[width=0.99\linewidth]{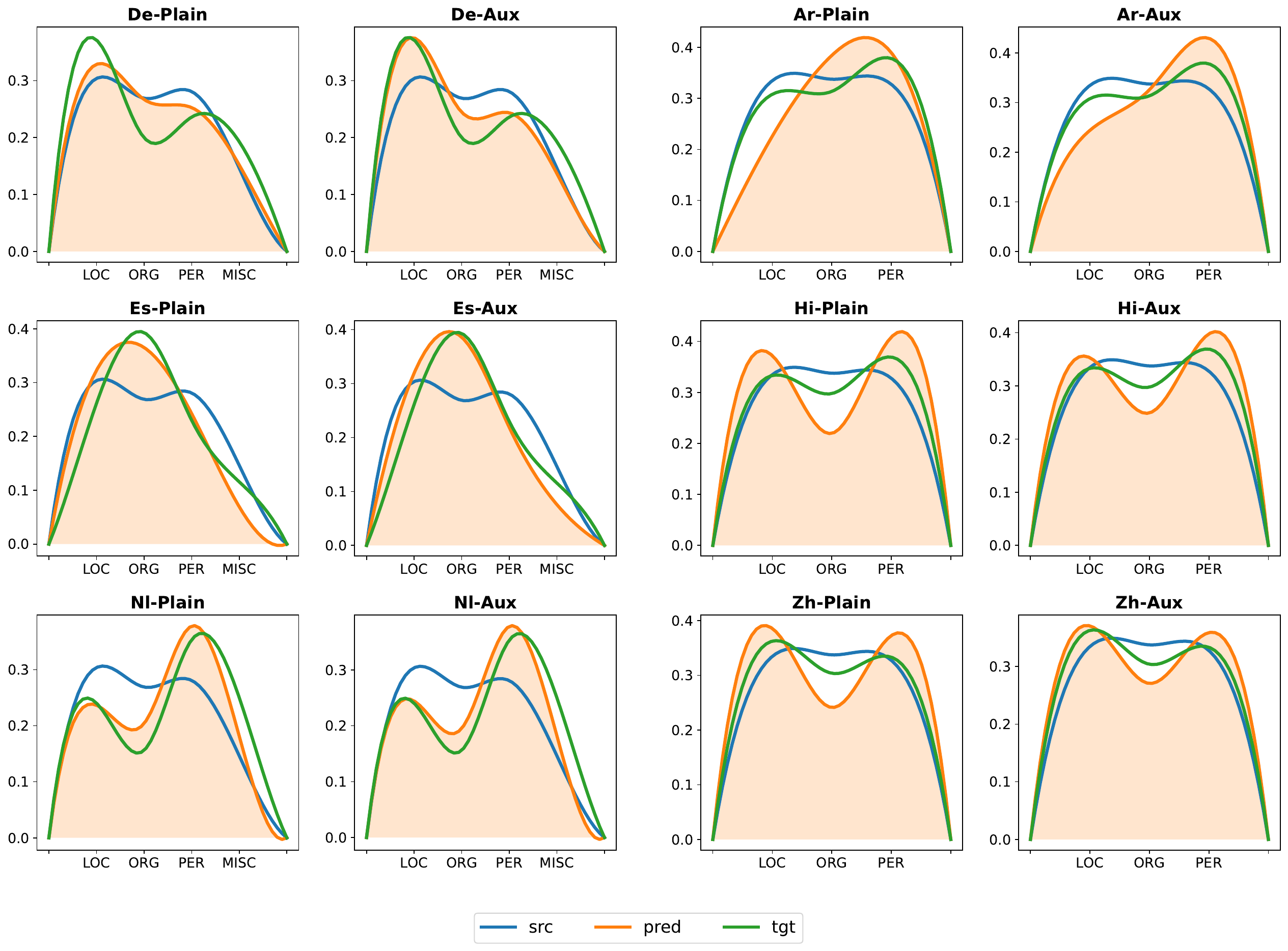}
    \caption{Distributions of entity types across all target languages (German (De), Spanish (Es), Dutch (Nl), Arabic (Ar), Hindi (Hi) and Chinese (Zh)). The vertical axis is the percentage of the entity type. ``-Plain'' in the figure title means that the predicted distribution is obtained from the model trained solely with source language data, and ``-Aux'' means that the predicted distribution is obtained from the model trained with both source language data and the auxiliary task. We interpolate discrete statistical results and represent them with continuous curves for better visualization.}
    \label{fig:label-dstr-all}
\end{figure*}

In section 4.5, we compare entity distributions from different data sources to analyze the effectiveness of target language-specific features, with German (De) data as an example. Here we give the analysis of all target languages for complement, as shown in Figure~\ref{fig:label-dstr-all}. Same as section 4.5, the source language entity distributions (blue lines), target language entity distributions (green lines), and model-predicted entity distributions for the target language (orange lines) are considered. ``-Plain'' in the figure title means that the predicted distribution is obtained from the model trained solely with source language data, and ``-Aux'' means that the predicted distribution is obtained from the model trained with both source language data and the auxiliary task. The predicted distributions shift towards the target language distributions, and the entity recognition accuracy improves for a wide range of entity types after introducing target language-specific features by the auxiliary task.

\section{Influence of LLMs on Cross-lingual NER}
The Large Language Models (LLMs), typified by ChatGPT, have demonstrated excellent comprehension of various languages. In this section, we briefly analyze their influence on cross-lingual NER.

Several studies have investigated the in-context learning ability of LLMs. For example, [Wei \textit{et al}., 2023]\footnote{
Zero-Shot Information Extraction via Chatting with ChatGPT
} 
evaluates ChatGPT's performance on NER in zero-shot scenarios with a multi-turn question-answering strategy. Their experiments on MSRA (a Chinese dataset) and Conllpp (an English dataset) obtained F1 scores of 57.7 and 58.4, respectively.
[Lai \textit{et al}., 2023]\footnote{ChatGPT Beyond English: Towards a Comprehensive Evaluation of Large Language Models in Multilingual Learning} experiments on more languages. Table~\ref{tab:LLM_test_MultiLang} presents part of their results. We conducted sample testing on the datasets we used, as shown in Table~\ref{tab:LLM_test_Demo}. Given the aforementioned results, a direct application of LLMs' in-context learning ability for NER might not be suitable, regardless of the language. Table~\ref{tab:LLM_cases} enumerates some cases. A notable factor leading to suboptimal performance is the design of prompts, which crucially impacts the LLMs' ability to execute specified tasks correctly. While potential improvement exists, we did not take such options given the complexity of the prompt design and implementation.

Some other studies leverage LLMs for assistance. For example, [Ma \textit{et al}., 2023]\footnote{Large Language Model Is Not a Good Few-shot Information Extractor, but a Good Reranker for Hard Samples!} employs small models for easy samples and large models for hard samples, achieving good performance on few-shot NER tasks. [Zhou \textit{et al}., 2024]\footnote{UniversalNER: Targeted Distillation from Large Language Models for Open Named Entity Recognition} leverages ChatGPT to generate data to fine-tune relatively small models (e.g. LLaMA), achieving good performance on open NER tasks. The exploration of this paradigm holds promise. For cross-lingual NER, effective utilization of the extensive multilingual knowledge of LLMs for assistance is a direction we will further explore in the future.

\begin{table}[!t]
\centering
\resizebox{0.99\linewidth}{!}{
\begin{tabular}{@{}lcccccc@{}}
\toprule
            & English & German & Spanish & Dutch & Hindi & Chinese \\ \midrule
prompt\_en  & 37.2    & 37.1   & 34.7    & 35.7    & 27.3  & 18.8  \\
prompt\_spc & 37.2    & 32.8   & 33.2    & 37.5    & 26.1  & 19.8  \\ \bottomrule
\end{tabular}
}
\caption{Performance of the ChatGPT (gpt-3.5-turbo) on MultiCoNER datasets from earlier work. F1 scores on 6 languages are reported. ``prompt\_en'' and ``prompt\_spc'' mean using English and corresponding languages to describe prompts, respectively.}
\label{tab:LLM_test_MultiLang}
\end{table}

\begin{table}[!t]
\centering
\resizebox{0.99\linewidth}{!}{
\begin{tabular}{@{}lccclccc@{}}
\toprule
          & \multicolumn{3}{c}{CoNLL} &  & \multicolumn{3}{c}{WikiAnn} \\ \midrule
          & German  & Spanish & Dutch &  & Arabic  & Hindi  & Chinese  \\ \midrule
Precision & 18.8    & 48.8    & 27.4  &  & 39.8    & 68.9   & 45.0     \\
Recall    & 61.6    & 70.3    & 64.2  &  & 35.3    & 73.1   & 46.9     \\
F1        & 28.8    & 57.6    & 38.4  &  & 37.5    & 70.9   & 46.0     \\ \bottomrule
\end{tabular}
}
\caption{Performance of the ChatGPT (gpt-3.5-turbo) on CoNLL and WikiAnn datasets. We randomly sample 100 sentences for each test and give one example in the input prompts.}
\label{tab:LLM_test_Demo}
\end{table}

\begin{table*}[!t]
\centering
\resizebox{0.99\linewidth}{!}{
\begin{tabular}{@{}p{\linewidth}@{}}
\toprule
\textbf{Input}: List the person, organization, and location in the text: ``Schatzmeister Wiesinger:''    \\
\textbf{Output1}:  \\
Person: Wiesinger  \\
Organization: \textcolor{red}{unknown organization of which Wiesinger is a member}    \\
\textbf{Output2}:    \\
Person: Wiesinger    \\
Organization: No specific organization mentioned  \\
Location: No specific location mentioned    \\
\hdashline
\\

\textbf{Input}: \\
\textit{Task Description}: You are asked to label a given text with named entity labels. The named entity labels are PER (person), LOC (location), and ORG (organization). You may encounter multi-word entities, so make sure to label each word of the entity with the appropriate prefix (``B'' for the first word of the entity, ``I'' for any non-initial word of the entity). For words that are not part of any named entity, you should return ``O''. \\
\textcolor{blue}{\textit{Note}: Your output format should be a list of tuples, where each tuple consists of a word from the input text and its corresponding named entity label.} \\
\textit{Input}: {[}``Schatzmeister'', ``Wiesinger'', ``:''{]}    \\
\textbf{Output}: \\
{[}(``Schatzmeister'', ``O''), (``Wiesinger'', ``PER''), (``:'', ``O''){]} \\
\hdashline
\\

\textbf{Input}:
List the person, location, and organization in the text. \textcolor{blue}{For example:}\\
\textcolor{blue}{Text:}\\
\textcolor{blue}{Am linken Bildrand ist Kronprinz Rupprecht von Bayern zu sehen.}\\
\textcolor{blue}{Output:}\\
\textcolor{blue}{location: Bayern; organization: None; person: Rupprecht von Bayern}\\
Text: Übersicht kostenfreie Downloads -Online-Umfrage Als Ergänzung zur internationalen Erhebung des Religionsmonitors bietet die Bertelsmann Stiftung eine Online-Umfrage in deutscher, englischer, spanischer und türkischer Sprache an.\\
\textbf{Output}: 
person: None; location: None; organization: Bertelsmann Stiftung\\
\hdashline
\\

\textbf{Input}: List the person, location, and organization in the text: Para poder ingresar a la OMC , China necesita cerrar acuerdos bilaterales con México , Ecuador , Costa Rica , Guatemala y Suiza . \\
\textbf{Output}:\\
Person: No specific person is mentioned in the text. \\
Location: \textcolor{red}{México, Ecuador, Costa Rica, Guatemala, Suiza} \\
Organization: OMC \\
\textbf{Input}: Why México, Ecuador, Costa Rica, Guatemala, and Suiza are locations?\\
\textbf{Output}: In this context, México, Ecuador, Costa Rica, Guatemala, and Suiza are locations \textcolor{red}{because they represent the countries with which China needs to close bilateral agreements to be able to join the World Trade Organization (WTO)}, which is an international organization that regulates trade between its member countries. Therefore, in this case, these countries are being referred to as locations where bilateral agreements need to be made.\\
\textbf{Input}: But I think México, Ecuador, Costa Rica, Guatemala, and Suiza are organizations in the context.\\
\textbf{Output}: You are correct, l apologize for my previous response. In the given context, México, Ecuador, Costa Rica, Guatemala, and Suiz should be classified as organizations \textcolor{red}{because they represent the countries with which China needs to make bilateral agreements to enter the World Trade Organization (WTO)}, an international organization that regulates trade between its member countries. Thank you for bringing this to my attention.\\
\\

\bottomrule
\end{tabular}
}
\caption{
Input-output cases of chatting with ChatGPT for NER. Four cases are presented. 
As shown in the first case, we discover that ChatGPT occasionally generates content beyond the provided text. Without constraining the randomness of its output, this issue may be handled through re-questioning.
In the second and third cases, we attempt to standardize the output format through text descriptions (referencing [Lai \textit{et al}., 2023]) or examples to facilitate the subsequent processing.
The last case shows an interesting conversation with ChatGPT. The model adjusts its conclusions under our inducement, but it gives the same explanation for the different conclusions.
}
\label{tab:LLM_cases}
\end{table*}

\end{document}